\pdfoutput=1

\documentclass[11pt]{article}

\usepackage{acl}

\usepackage{times}
\usepackage{latexsym}

\usepackage[T1]{fontenc}

\usepackage[utf8]{inputenc}

\usepackage{microtype}

\usepackage{inconsolata}

\usepackage{graphicx}

\usepackage{algorithm}
\usepackage{algorithmic}
\usepackage{amsmath}
\usepackage{booktabs}
\usepackage{multirow}
\usepackage{amsfonts}
\usepackage{mdframed}
\usepackage{tabularx,array,booktabs}

\title{Self-Training Meets Consistency: Improving LLMs' Reasoning with Consistency-Driven Rationale Evaluation}

\author{Jaehyeok Lee\\
  Sungkyunkwan University \\
  Suwon, South Korea \\
  \texttt{hjl8708@skku.edu} \\
  \And
  Keisuke Sakaguchi \\
  Tohoku University \\
  Sendai, Japan \\
  \texttt{keisuke.sakaguchi@tohoku.ac.jp} 
  \And
  JinYeong Bak \\
  Sungkyunkwan University \\
  Suwon, South Korea \\
  \texttt{jy.bak@skku.edu} 
  \\}

\author{
 \textbf{Jaehyeok Lee$^{1}$},
 \textbf{Keisuke Sakaguchi$^{2^*}$},
 \textbf{JinYeong Bak$^{1^*}$}
\\
$^1$Sungkyunkwan University, Suwon, South Korea \\
  $^2$Tohoku University, Sendai, Japan \\
  \texttt{hjl8708@skku.edu}, \texttt{keisuke.sakaguchi@tohoku.ac.jp}, \texttt{jy.bak@skku.edu} \\
}

\newcommand{\modelname}{CREST}
\newcommand{\modelfullname}{Consistency-driven Rationale Evaluation for Self-Training}

\newcommand{\hlinemargin}{
\vskip1mm
\hrule
\vskip1mm
}

\begin{document}

\maketitle

\begingroup\def\thefootnote{*}\footnotetext{Corresponding authors}\endgroup
\renewcommand{\thefootnote}{\arabic{footnote}}

\begin{abstract}
Self-training approach for large language models (LLMs) improves reasoning abilities by training the models on their self-generated rationales.
Previous approaches have labeled rationales that produce correct answers for a given question as appropriate for training.
However, a single measure risks misjudging rationale quality, leading the models to learn flawed reasoning patterns.
To address this issue, we propose CREST (Consistency-driven Rationale Evaluation for Self-Training), a self-training framework that further evaluates each rationale through follow-up questions and leverages this evaluation to guide its training.
Specifically, we introduce two methods: (1) filtering out rationales that frequently result in incorrect answers on follow-up questions and (2) preference learning based on mixed preferences from rationale evaluation results of both original and follow-up questions.
Experiments on three question-answering datasets using open LLMs show that CREST not only improves the logical robustness and correctness of rationales but also improves reasoning abilities compared to previous self-training approaches.\footnote{Code: \href{https://github.com/JaehyeokLee-119/CREST}{https://github.com/JaehyeokLee-119/CREST}}
\end{abstract}

\section{Introduction}
Large language models (LLMs) can enhance multi-step reasoning abilities by generating intermediate reasoning steps (i.e., rationale) before arriving at an answer~\cite{wei2022chain}. 
Training LLMs on high-quality rationales has been shown to improve their reasoning capabilities~\cite{chungJMLR:v25:23-0870scaling, liu2023logicot, shridhar-etal-2023-distilling}.
Therefore, collecting high-quality rationales is becoming increasingly important for training the reasoning abilities of LLMs.
However, due to the high cost associated with collecting high-quality rationales, self-training approaches have emerged, focusing on training LLMs using self-generated rationales~\cite{STaR}.
\begin{figure}[t!]
    \centering
    \includegraphics[width=\linewidth]{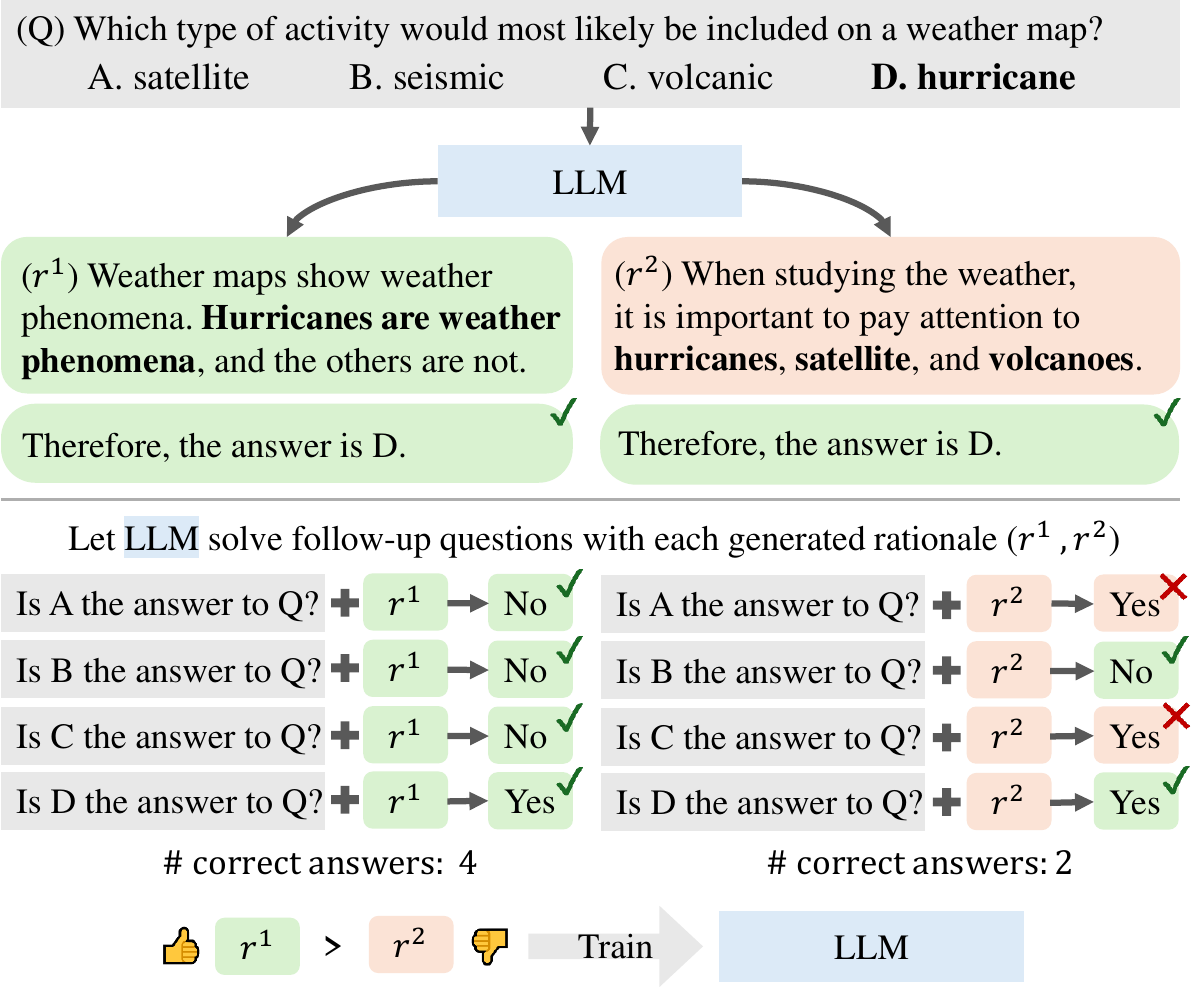}
    \caption{
    An example of rationale generation and evaluation in \modelname: An LLM generates two rationales ($r^1$, $r^2$) and answer predictions to solve question Q.
    Even though $r^2$ lacks focus and clear support for the answer, previous approaches evaluate both $r^1$ and $r^2$ as equally right. Through a more fine-grained evaluation using follow-up questions, we can identify the better rationale, $r^1$, which leads to more consistent predictions across all questions.
    }
    \label{fig:introduction}
\end{figure}

In self-training approaches, accurately evaluating the quality of generated rationales is essential. Previous studies have evaluated rationale quality by examining whether the generated rationales lead to the correct answer to a given question~\cite{STaR, hoffman2023training, feng2024improving, hosseini2024vstar, singh2024beyond}.
However, using the correctness of a single prediction is unstable, as LLMs can reach correct answers through inappropriate reasoning steps~\cite{bao2024llmschainofthoughtnoncausalreasoners}.
Figure \ref{fig:introduction} shows an example of two generated rationales, $r^1$ and $r^2$. Despite $r^2$ shows incomplete reasoning, previous approaches would consider both rationales equally appropriate since they both lead to the correct answer for Q.
Training models on such inappropriate rationales can cause them to learn flawed reasoning patterns.

To address this problem, we propose \modelname\ (\modelfullname), a novel framework for LLM self-training. 
The core idea of \modelname\ is to further evaluate rationales using follow-up questions that ask whether each answer option in the original question is correct or not.
We first generate diverse rationales using temperature sampling and evaluate them with an LLM as shown in Figure \ref{fig:introduction}. 
Subsequently, we train the LLM on these rationales, rewarding rationales that lead to more consistent predictions (i.e., $r^1$) and penalizing those that lead to less consistent predictions (i.e., $r^2$).
To achieve this, we propose two methods: rationale filtering and preference learning.
In rationale filtering, we remove rationales that lead to incorrect answers in more than a certain number of follow-up questions during the supervised fine-tuning process.
In preference learning, we train the model on mixed preferences from results of both original and follow-up questions, to favor rationales that result in correct answers in a greater number of follow-up questions.

We conduct experiments on three natural language reasoning question-answering datasets, including ReClor~\cite{yu2020reclor}, ARC~\cite{Clark2018ThinkYH}, and CSQA~\cite{talmor-etal-2019-commonsenseqa}.
We compare \modelname\ to other self-training approaches using Llama 3 model~\cite{llama3modelcard} and Gemma model~\cite{gemmateam2024gemmaopenmodelsbased}. 
Our findings show that \modelname\ can train an LLM to generate more correct and robust rationales, improving its reasoning performance.
Our contributions are as follows:
\begin{itemize}
    \item We introduce consistency-driven rationale evaluation, which further evaluates generated rationales using follow-up questions that ask whether each answer option in the original question is correct or not.
    \item We propose \modelname, which evaluates generated rationales via consistency-driven rationale evaluation and uses the evaluation results to train an LLM through two methods: rationale filtering and preference learning using mixed preferences derived from original and follow-up question evaluations.
    \item We conduct experiments and analyses with open LLMs such as Llama 3 model and Gemma model on three question-answering datasets. The results show that \modelname\ generates more robust and correct rationales and improves reasoning ability compared to other self-training approaches.
\end{itemize}

\section{Related Work}
\begin{figure*}[t!]
    \centering
    \includegraphics[width=\linewidth]{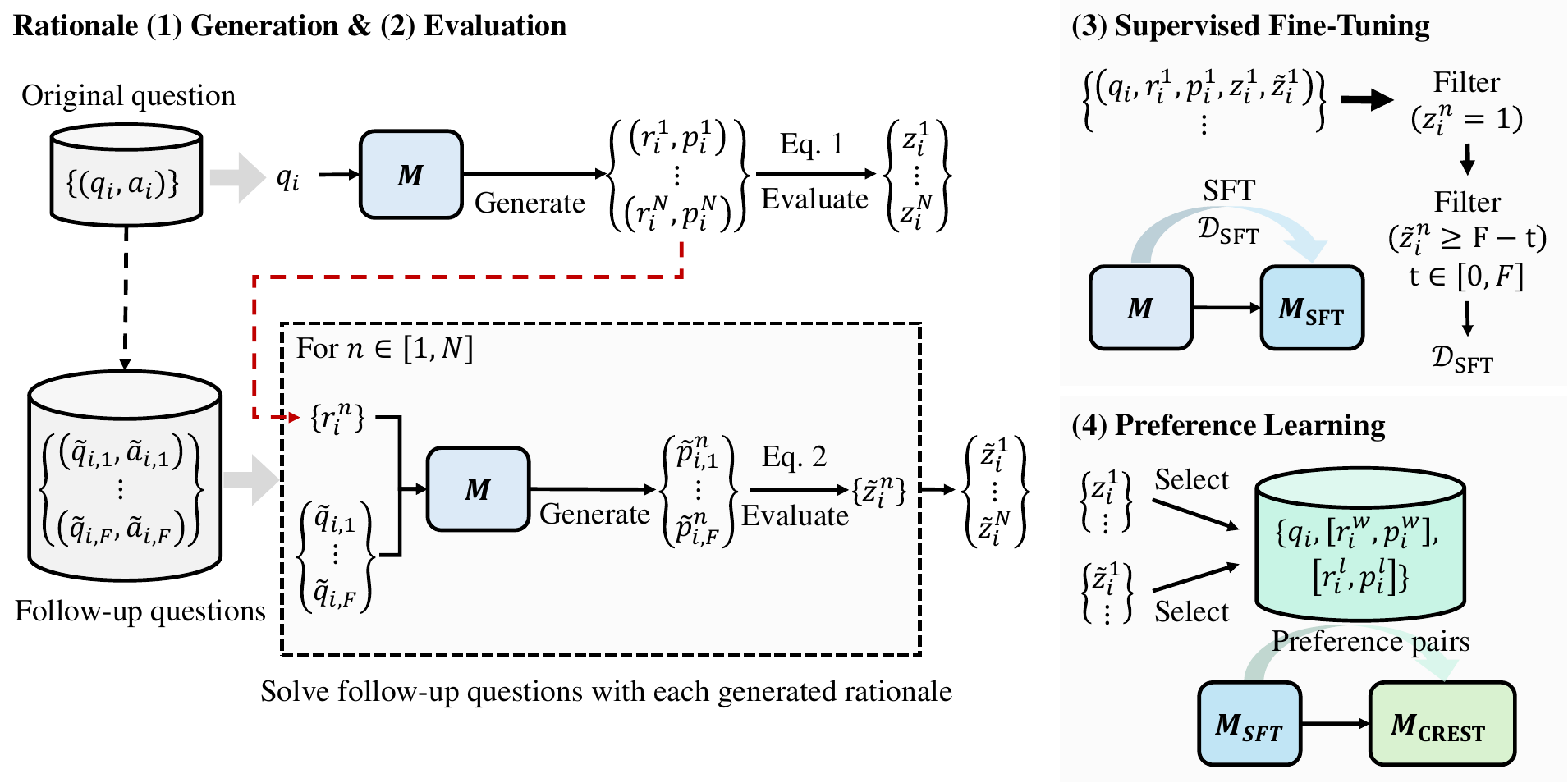}
    \caption{
    Overview of \modelname. In Rationale Generation (1), given a question $q_i$ and an answer $a_i$, an initial LLM $\textbf{\textit{M}}$ generates $N$ rationales and answer predictions $(r_i, p_i)$ to solve $q_i$, and then solves follow-up questions $\tilde{q}_{i,f}$ using each rationale $r^n_i$, resulting in $\tilde{p}^n_{i,f}$. Next, in Rationale Evaluation (2), we assign rewards $z$ and $\tilde{z}$ to each rationale based on the correctness of the predictions as shown in Eq. \ref{eq:z} and Eq. \ref{eq:tilde_z}. In Supervised Fine-Tuning (3), we train $\textbf{\textit{M}}$ on the rationales filtered by $z$ and $\tilde{z}$ with a tolerance term $t$, resulting in $\textbf{\textit{M}}_\textbf{SFT}$. Finally, in Preference Learning (4), we build preference pairs based on $z$ and $\tilde{z}$, and train $\textbf{\textit{M}}_\textbf{SFT}$ on them, resulting in $\textbf{\textit{M}}_\textbf{\modelname}$.
    }
    \label{fig:rationale_gen_and_pred}
\end{figure*}

\subsection{Self-Training Approaches}
Chain-of-Thought (CoT) approach demonstrates that generating a step-by-step reasoning path before the final prediction enhances an LLM's reasoning abilities~\cite{wei2022chain}. 
Training LLMs on rationale data generated by humans~\cite{chungJMLR:v25:23-0870scaling} or advanced models like GPT-4 further enhances reasoning abilities~\cite{liu2023logicot}.
However, since high-quality rationale data is expensive to obtain, a number of approaches focus on training language models using self-generated rationales. STaR~\cite{STaR}, an early type of self-training approach, trains the language model by selecting the correct rationales based on binary feedback regarding the correctness of the answers generated by these rationales. 
RFT~\cite{yuan2023scaling} enhances supervised data by generating and collecting diverse correct reasoning paths, focusing on mathematical reasoning. 
Other approaches, such as V-STaR, Iterative RPO, and Self-motivated Learning, also utilize incorrect rationales~\cite{feng2024improving, hosseini2024vstar, pang2024iterative} and adopt preference learning techniques, such as Proximal Policy Optimization (PPO)~\cite{schulman2017proximal} and Direct Preference Optimization (DPO)~\cite{rafailov2023direct}. 
Self-Explore~\cite{hwang2024selfexplore} provides fine-grained rewards by identifying incorrect steps within the rationales.
\citet{wei-jie-etal-2024-self} proposes a self-training framework that exposes a model to each question multiple times using temperature sampling, thereby assessing the model's confidence in the given question.
\modelname\ provides fine-grained rewards through evaluating a self-training rationale multiple times using follow-up questions augmented from the original dataset, emphasizing the rationale's ability to consistently lead to correct answers.

\subsection{Reasoning with Consistency} 
Consistency is the ability to make consistent decisions in semantically equivalent contexts~\cite{elazar-etal-2021-measuring}. It is a desirable property of logically valid machine learning systems~\cite{chen2024two} and an important characteristic for a model to be considered trustworthy~\cite{jang-etal-2022-becel}. 
As larger language models emerge that exceed human performance in many tasks, consistency is receiving increased attention due to its role in evaluating inference validity, even in models that outperform humans~\cite{Fluri2024Evaluating}.
To evaluate a model's consistency, follow-up questions generated from existing questions are commonly used~\cite{ribeiro-etal-2019-red, elazar-etal-2021-measuring, jang-etal-2022-becel, chen2024two, zheng2024trustscore, chen2024consistentnaturallanguageexplanationsexplanationconsistency}. 
Several techniques have been developed to create these follow-up questions, including generating semantically identical texts by paraphrasing the original input texts~\cite{elazar-etal-2021-measuring}, crafting logically equivalent questions~\cite{jang-etal-2022-becel}, and developing questions that investigate the implications of the model's answers~\cite{ribeiro-etal-2019-red}. 
Two main approaches have been proposed to enhance both the consistency and task performance of models: designing models specifically to reduce inconsistency~\cite{kassner-etal-2021-beliefbank, kassner-etal-2023-language}, and synthesizing consistent data to train models~\cite{alberti-etal-2019-synthetic, asai-hajishirzi-2020-logic, elazar-etal-2021-measuring}.
\modelname\ evaluates rationales that correspond to the reasoning process with augmented questions and trains an LLM to prefer those that consistently lead to correct answers.

\section{\modelfullname}

This section describes our approach, \modelfullname\ (\modelname) which trains reasoning abilities through consistency-driven rationale evaluation with follow-up questions.

\subsection{Notation}
We have a pretrained large language model $\textbf{\textit{M}}$ and an original dataset of questions $q$ with answers $a$, represented as $\mathcal{D}=\{(q_i, a_i)\}^D_{i=1}$. Each question has $F$ answer choices. To solve $q$, \textbf{\textit{M}} sequentially generates a rationale $r$, corresponding to intermediate reasoning steps, and an answer prediction $p$, where $r$ leads to $p$.

\subsection{\modelname}
The whole framework of \modelname\ consists of four stages.
Figure \ref{fig:rationale_gen_and_pred} outlines the overview of \modelname.
\begin{itemize}
    \item \textbf{Rationale Generation} We generate $N$ diverse rationales $r^n_i$ for each question $q_i$ and the corresponding answer predictions $p^n_i$ using \textbf{\textit{M}}, where $n\in[1,N]$.
    \item \textbf{Rationale Evaluation} We compare $p^n_i$ with $a_i$ to assign a reward $z^n_i$ to $r^n_i$ based on the correctness of the prediction.
    Subsequently, we generate multiple follow-up questions $\tilde{q}_{i,f}$ from $q_i$ and further evaluate $r^n_i$ using these follow-up questions. We assign an additional reward $\tilde{z}^n_i$ to $r^n_i$ based on how many $\tilde{q}_{i,f}$ are answered correctly.
    \item \textbf{Supervised Fine-Tuning} We train \textbf{\textit{M}} through supervised fine-tuning to create $\textbf{\textit{M}}_\textbf{SFT}$ using the generated rationales filtered based on the evaluation results.
    \item \textbf{Preference Learning} We train $\textbf{\textit{M}}_\textbf{SFT}$ using a preference learning algorithm according to the preferences indicated by the evaluation results, resulting in $\textbf{\textit{M}}_\textbf{\modelname}$.
\end{itemize}

\subsection{Rationale Generation}
Initially, we generate diverse rationales and the corresponding answer predictions for a given original question $q_i$ with \textbf{\textit{M}}. Specifically, \textbf{\textit{M}} generates $N$ rationales $r^n_i$ as follows: $ r^n_i \leftarrow \textbf{\textit{M}}(q_i)$,
where $r^n_i$ represents the $n^\text{th}$ rationale generated for the $i^\text{th}$ question.
Subsequently, \textbf{\textit{M}} derives answer predictions $p^n_i$ for $q_i$ from generated rationales $r^n_i$, as follows: $p^n_i \leftarrow \textbf{\textit{M}}(q_i, r^n_i)$.

\subsection{Rationale Evaluation}
We evaluate the rationale through a two-step process. Firstly, similar to previous studies~\cite{STaR, yuan2023scaling, hosseini2024vstar, feng2024improving, pang2024iterative}, we compare the ground truth answer $a_i$ for $q_i$ with the predicted answer $p^n_i$ derived from $r^n_i$. 
Secondly, we further assess the rationales through $F$ follow-up questions which are generated from the original question $q_i$.

In the first step, we assign a binary reward $z^n_i$ of either 0 or 1 to each rationale based on whether $p^n_i$ matches $a_i$ as follows: 
\begin{equation}
z^n_i=\mathbf{1}(p^{n}_{i} = a_{i})
\label{eq:z}
\end{equation} 
Assuming that rationales leading to the correct answer are of higher quality than those that do not, as suggested by~\citet{STaR}, this evaluation directly measures the quality of rationales.

In the second step, we evaluate the rationales using $F$ follow-up questions $\{(\tilde{q}_{i,1},\tilde{a}_{i,1}), ...,(\tilde{q}_{i,F},\tilde{a}_{i,F})\}$ generated from $q_i$, where $\tilde{a}_{i,f}$ is the ground truth answer for the $f^{\text{th}}$ follow-up question corresponding to $q_i$.
We then evaluate the rationales on all $F$ follow-up questions: $\tilde{p}^{n}_{i,f} \leftarrow \textbf{\textit{M}}(\tilde{q}_{i,f}, r^n_i)$, where $\tilde{q}_{i,f}$ is $f^\text{th}$ follow-up question for $q_i$.

We assign an additional reward $\tilde{z}$ to each rationale based on the number of correctly solved follow-up questions as follows:
\begin{equation}
  \tilde{z}^n_i = \sum_{f=1}^{F} \mathbf{1}(\tilde{p}^{n}_{i,f} = \tilde{a}_{i,f})
\label{eq:tilde_z}
\end{equation}

To generate follow-up questions that are closely related to the problem-solving process of each question in $\mathcal{D}$, we utilize the characteristics of multiple-choice questions: the solving process involves not only identifying the correct answer but also eliminating the incorrect options.
We design each follow-up question to ask whether each of the answer options in the original question is correct or not. 
This type of follow-up question is used to evaluate the robustness of reasoning ability in multiple-choice question-answering datasets \cite{wang2024answersreviewingrationalitymultiple}.
Figure \ref{fig:introduction} shows an example of the follow-up questions and the evaluation.

\subsection{Supervised Fine-Tuning}
\label{rationale_filtering}
After evaluating the rationales, we use $z$ and $\tilde{z}$ as filters to select the rationales for training \textbf{\textit{M}} and produce $\textbf{\textit{M}}_\textbf{SFT}$ through supervised fine-tuning (SFT).
Intuitively, the best rationales for $q_i$ from the previous stage are those that lead to the correct answers to $q_i$ and all $F$ follow-up questions, indicated by $z^n_i=1$ and $\tilde{z}^n_i=F$.
However, simply removing rationales that lead to incorrect answers for any of the follow-up questions might drastically reduce the number of rationales available for training. Therefore, we also include some sub-optimal rationales with a tolerance term $t$ that satisfies $t\in [0,F]$. Consequently, the dataset $\mathcal{D_{\text{SFT}}}$ used to train \textbf{\textit{M}} in the SFT stage is represented as follows:
\begin{alignat}{2}
    \label{eq:tolerance_filtering}
        \mathcal{D_{\text{SFT}}} = &\{q_i, r^n_i, a_i | \\
        &(n, i) \in \{(n, i)|z^n_i=1, \tilde{z}^n_i \geq F-t\}\} \nonumber
\end{alignat}
The training objective for this stage aligns with that used during pretraining, specifically employing an auto-regressive language modeling objective or next-token prediction \cite{radford2018improving}. 
We calculate the loss exclusively for the output section (i.e., $r$ and $a$).

\subsection{Preference Learning} 
We further train $\textbf{\textit{M}}_\textbf{SFT}$ by exploiting preferences between rationales to enhance its reasoning ability.
To achieve this, we construct preference pairs and fine-tune $\textbf{\textit{M}}_\textbf{SFT}$ using offline preference learning methods, such as Direct Preference Optimization (DPO)~\cite{rafailov2023direct}.

\subsubsection{Preference Pair Dataset Construction}
We construct the preference pair dataset $P_{\textit{total}}$ for preference learning by first creating two sets of preference pairs $P_{z}$ and $P_{\tilde{z}}$, which represent rationale preferences based on the rewards $z$ and $\tilde{z}$, respectively. $P_{\textit{total}}$ is then formed by randomly sampling pairs from these two sets.

To construct $P_{z}$ and $P_{\tilde{z}}$, we generate preference pairs in which rationales with higher rewards $r^w$ are preferred over those with lower rewards $r^l$, based on $z$ and $\tilde{z}$, respectively. Each preference pair consists of a question $q$, two generated rationales, and their corresponding predictions $p^w$ and $p^l$: $(q, r^w, p^w, r^l, p^l$).
Algorithm \ref{algorithm:P_oq} outlines the detailed procedure for generating $P_{z}$ and $P_{\tilde{z}}$.

Then, we construct $P_{\textit{total}}$ by sampling pairs from $P_{z}$ and $P_{\tilde{z}}$ with a weighting factor $\lambda$, which controls the relative contribution of rationale preferences derived from $z$ and $\tilde{z}$ during preference learning.
The parameter $\lambda$ satisfies $\lambda \in [0,1]$, ensuring that the proportion of $P_{\tilde{z}}$ in $P_{\textit{total}}$ is $\lambda$.
For instance, if a total of 10,000 pairs are used for preference learning and $\lambda=0.4$, $P_{\textit{total}}$ would consist of 4,000 randomly selected pairs from $P_{\tilde{z}}$ and 6,000 randomly selected pairs from $P_z$. The total number of pairs used for preference learning is determined by the maximum number of training steps multiplied by the batch size.

\subsubsection{Training}
We train $\textbf{\textit{M}}_\textbf{SFT}$ on the preference pairs $P_{\textit{total}}$ using DPO, resulting in $\textbf{\textit{M}}_\textbf{\modelname}$.
Given the preference pairs $P_{\textit{total}}$, the objective of this stage is to increase the log-likelihood of preferred outputs over dispreferred ones:
\begin{alignat}{2}
        &\mathcal{L_{\text{DPO}}}=-\mathbb{E}_{(r^w_i,p^w_i,r^l_i,p^l_i,q_i)\sim P_{\textit{total}}} 
        \\&\Big[\log\sigma\Big({\hat{r}_\theta(q_i,r^w_i,p^w_i)}-\hat{r}_\theta(q_i,r^l_i,p^l_i)\Big)\Big] \nonumber
        \\&\hat{r}_\theta(q,r,p)=\beta\log{\frac{\pi_\theta(r, p | q)}{\pi_{\textit{ref}}(r, p | q)}}
\end{alignat}
where $\pi_\theta(r,p|q)$ and $\pi_{\textit{ref}}(r,p|q)$ represent the probability of outputs $r$ and $p$ given input $q$ under the current policy parameterized by $\theta$ and a reference policy $\pi_{\textit{ref}}$, respectively. 
Initially, both $\pi_\theta$ and $\pi_{\textit{ref}}$ are initialized as $\textbf{\textit{M}}_\textbf{SFT}$, and they are updated each epoch.
$\pi_{\textit{ref}}$ is used to minimize distribution shift from the true reference distribution and is typically initialized through supervised fine-tuning on preferred outputs.
$\beta$ controls the deviation from the reference policy.

\section{Experiments}
\begin{table*}[t]
\centering
{
\begin{tabular}{clrrrrrr}
\toprule
& \multicolumn{1}{l}{Base model}        & \multicolumn{3}{c}{Llama 3 8B}                    & \multicolumn{3}{c}{Gemma 7B}                     \\ \midrule 
\multicolumn{1}{c}{Approach} & \multicolumn{1}{l}{Model}             & \multicolumn{1}{c}{ReClor} & \multicolumn{1}{c}{ARC-C}  & \multicolumn{1}{c}{CSQA}  & \multicolumn{1}{c}{ReClor}    & \multicolumn{1}{c}{ARC-C} & \multicolumn{1}{c}{CSQA}  \\ \midrule
\multirow{2}{*}{}& Zero-shot                                      & 52.10          & 69.28          & 53.89          & 53.60          & 77.47          & 65.68          \\
& Few-shot                                       & 55.30          & 77.21          & 70.76          & 58.70          & 83.11          & 75.02          \\ \midrule
\multirow{5}{*}{Self-training}& STaR                                           & 58.60          & 77.99          & 76.17          & 58.40          & 82.34          & 77.56          \\
& RFT                                            & 64.40          & 80.72          & 78.54          & 66.90          & 83.36          & 80.02          \\
 & Self-motivated Learning                        & 67.80          & 80.03          & 80.34          & 68.20          & 83.53          & 80.59          \\ 
& $M_\text{SFT}$                                 & 66.10          & 81.40          & 79.36          & 67.90          & 84.22          & 80.51          \\
& $M_\text{\modelname}$                          & \textbf{69.50}          & \textbf{81.91} & \textbf{81.41} & \textbf{70.00}          & \textbf{84.47}          & \textbf{80.67}          \\ \midrule
\multirow{2}{*}{\begin{tabular}[c]{@{}c@{}}Direct\\ fine-tuning\end{tabular}}  & Fine-tune (Label)                              & 77.40          & 80.80          & 80.18          & 81.90          & 85.58          & 84.44          \\
& $\text{Fine-tune (Label)}_{\text{\modelname}}$ & \textbf{79.30} & \textbf{81.23}          & \textbf{81.24}          & \textbf{83.70} & \textbf{87.20} & \textbf{84.85} \\ \bottomrule
\end{tabular}
}
\caption{Accuracy of various models across three reasoning datasets with Llama 3 8B and Gemma 7B model. 
ARC-C denotes the challenge set in the ARC test set. 
\modelname\ consistently improves accuracy across all three datasets.
}
\label{table:main_result}
\end{table*}

This section describes the experiments and results of \modelname\ compared to other self-training approaches. 
First, we introduce the three datasets used for model training and testing. Next, we present the experimental setup, including the base LLM, key hyperparameters, and performance metrics. We also introduce the baseline approaches used for comparison, and finally, we present the results of the experiments.

\subsection{Experimental Settings}
\paragraph{Datasets}
We evaluate \modelname\ on three English natural language reasoning multiple-choice QA datasets: ReClor~\cite{yu2020reclor}, ARC~\cite{Clark2018ThinkYH}, CSQA~\cite{talmor-etal-2019-commonsenseqa}. ReClor comprises logical reasoning problems derived from American graduate school entrance exams and their preparatory materials. 
The ReClor test set is divided into an Easy set, which consists of biased data points, and a Hard set, which includes the remaining data points.
ARC is sourced from grade-school science assessments for students of various grades. The questions are categorized into two sets: an Easy set and a Challenge set. In our experiments, we only test on the Challenge set, as in previous studies \cite{huang-etal-2023-large, pang2024iterative}. 
CSQA consists of short questions that require common sense reasoning, built upon ConceptNet~\cite{Speer_Chin_Havasi_2017}.
\paragraph{Models}
We conduct our experiments using the Llama 3 8B model\footnote{\url{https://huggingface.co/meta-llama/Meta-Llama-3-8B}}~\cite{llama3modelcard} and the Gemma 7B model\footnote{\url{https://huggingface.co/google/gemma-7b}}~\cite{gemmateam2024gemmaopenmodelsbased} from \textit{HuggingFace}~\cite{wolf-etal-2020-transformers}, training them with Low-Rank Adaptation (LoRA)~\cite{hu2022lora}.
\paragraph{Implementation Details}
We generate rationales with temperature sampling with the following parameters: $T$=0.8, $TopP$=0.95, $N$=16, and $\text{max\_new\_tokens}$=512, then use greedy decoding for answer prediction. 
For supervised fine-tuning, we use epoch=6, batch size=32 and conduct learning rate search between \{$5e-6$, $5e-3$\}.
For preference learning, we use $\beta$=0.1, epoch=4, batch size=8, and search max number of steps among \{3000, 5000\} and conduct learning rate search between \{$5e-7$, $5e-5$\} for all models.
The input and output prompt templates for model evaluation are illustrated in Figures \ref{fig:rationale_generation_template} and \ref{fig:without_CoT_template}.
For more details about the prompts used in this study, please refer to Appendix \ref{Appendix_prompts}. 

\subsection{Baselines}

\begin{itemize}
\item\textbf{Fine-tune (Label)} involves directly fine-tuning the base model on ground truth labels using a negative log-likelihood loss term, without relying on any generated rationales.

\item\textbf{STaR~\cite{STaR}} is an early approach for generating, filtering, and learning rationales using a generative language model. It generates a rationale for each question and trains the language model on rationales that lead to correct predictions. Additionally, STaR introduces a rationalization process that provides hints when the initial rationale fails to produce a correct prediction.

\item\textbf{RFT~\cite{yuan2023scaling}} stands for Rejection Sampling Fine-Tuning. RFT generates diverse rationales with a non-zero temperature and selects rationales to train based on binary feedback on the correctness of the final prediction. Unlike STaR, RFT does not have a rationalization process. In our experiments, $\textbf{\textit{M}}_\textbf{SFT}$ with maximum tolerance corresponds to RFT.

\item\textbf{Self-motivated Learning~\cite{feng2024improving}} exploits the inherent preference between correct rationales and incorrect rationales. 
It first trains a base model on generated and filtered rationales through supervised fine-tuning.
It trains a reward model that assigns higher rewards to correct rationales than to incorrect ones. 
This reward model is then used to improve the reasoning performance of a supervised fine-tuned model through reinforcement learning using Proximal Policy Optimization (PPO)~\cite{schulman2017proximal}. 
\end{itemize}

\subsection{\modelname}

\begin{itemize}
\item\textbf{$\textit{M}_\text{SFT}$} is supervised fine-tuned on filtered rationales from the base model. The performance difference between this model and RFT demonstrates the effect of the rationale filtering process.

\item\textbf{$\textit{M}_\text{CREST}$ \& $\text{Fine-tune (Label)}_\text{\modelname}$} are models trained using preference learning in \modelname, based on  $\textbf{\textit{M}}_\textbf{SFT}$ and $\text{Fine-tune (Label)}$, respectively.
To evaluate the effectiveness of preference learning with $P_{\textit{total}}$, we apply it to two models: $\textbf{\textit{M}}_\textbf{SFT}$, a model fine-tuned on filtered rationales, and $\text{Fine-tune (Label)}$, a model fine-tuned directly on ground truth labels.
For details on the prompt templates used to train $\text{Fine-tune (Label)}_\text{\modelname}$,
please refer to Appendix \ref{sec:finetune_crest_template}.
The resulting models, named $\textbf{\textit{M}}_\textbf{\modelname}$ and $\text{Fine-tune (Label)}_\text{\modelname}$, demonstrate how \modelname\ enhances reasoning performance through preference learning.
\end{itemize}

\begin{figure*}[t!]
\centering
    \includegraphics[width=\linewidth]{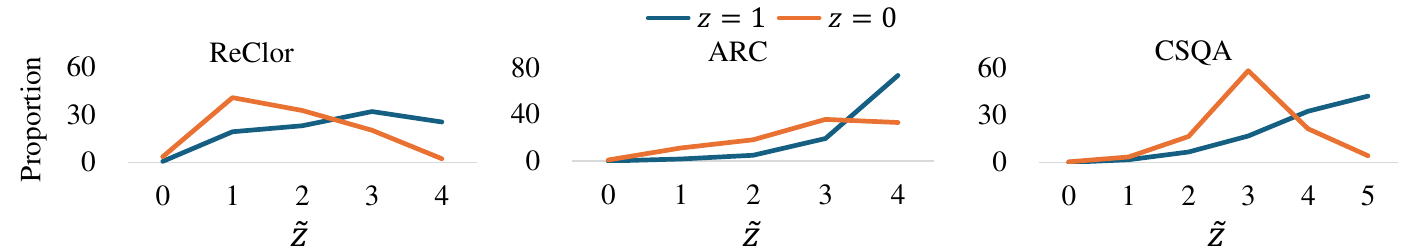}
    \caption{
    Distribution of rationale proportions based on $\tilde{z}$ for rationales with $z=1$ and $z=0$, respectively. 
    For example, among the generated rationales with $z=0$ for CSQA, approximately 60\% have $\tilde{z}=3$.
    Rationales with $z=0$ are relatively concentrated at lower $\tilde{z}$ values compared to those with $z=1$.
    This correlation between $z$ and $\tilde{z}$ suggests that $\tilde{z}$ reflects the quality of the rationale.
    }
    \label{fig:wrong_rationales_ratio}
\end{figure*}
\begin{figure*}[t!]
\centering
    \includegraphics[width=\linewidth]{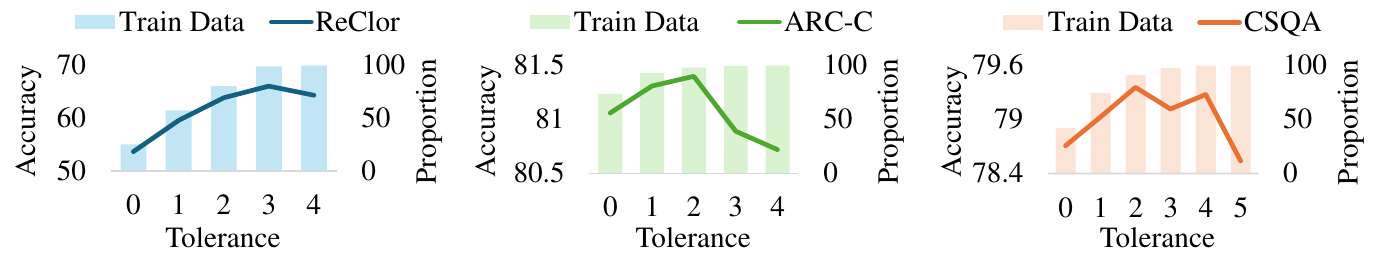}
    \caption{
    Proportion of rationale data used for training $\textbf{\textit{M}}_\textbf{SFT}$ and task performance on three datasets, according to tolerance $t$.
    The results suggest that while moderate tolerance $t$ improves performance, while overly high $t$ values can degrade it, indicating the importance of excluding less robust rationales from training.
    }
    \label{fig:tolerance_performance}
\end{figure*}

\subsection{Results}
As shown in Table \ref{table:main_result}, $\textbf{\textit{M}}_\textbf{\modelname}$ outperforms other self-training baselines across the three datasets. 
Both RFT and $\textbf{\textit{M}}_\textbf{SFT}$ are models trained through supervised fine-tuning on the base model, with the key difference being whether rationale filtering based on $\tilde{z}$ was applied. 
The result that $\textbf{\textit{M}}_\textbf{SFT}$ outperforms RFT across all three datasets demonstrates that rationale filtering based on $\tilde{z}$ consistently improves performance while reducing the amount of training data.
Comparing $\textbf{\textit{M}}_\textbf{SFT}$ with $\textbf{\textit{M}}_\textbf{\modelname}$, and Fine-tune (Label) with $\text{Fine-tune (Label)}_\text{\modelname}$, we can see that preference learning with pairwise preference datasets constructed using follow-up questions consistently enhances performance across all three datasets.

\section{Analysis}

In this section, we explore the effectiveness of consistency-driven evaluation and the impacts of rationale filtering and preference learning in \modelname\ on model performance, through analyses using the Llama 3 8B model as the base model.
Our analysis includes examining the correlation between $z$ and $\tilde{z}$ and conducting ablation studies on parameters such as $t$ and $\lambda$ to assess how the proposed methods in \modelname\ contribute to performance improvement.
To investigate the impact of preference learning with $P_{\tilde{z}}$, we create a model that trains $\textbf{\textit{M}}_\textbf{SFT}$ using preference learning with only $P_z$, which we refer to as 
$\textbf{\textit{M}}_{\textbf{SFT /w } \boldsymbol{P_{z}}}$, and compare it to $\textbf{\textit{M}}_\textbf{\modelname}$.

\subsection{Incorrect Rationales on Follow-up Questions}
To understand how evaluation through follow-up questions reflects the quality of rationales, we evaluate incorrect rationales ($z=0$) generated from train datasets on the follow-up questions, as shown in Figure \ref{fig:wrong_rationales_ratio}.
The incorrect rationales are less robust on follow-up questions compared to correct rationales $(z=1)$, especially incorrect rationales have a significantly lower rate of getting all follow-up questions correct. 
This correlation between $z$ and $\tilde{z}$ indicates that $\tilde{z}$ can reflect the quality of a rationale.

\subsection{Effect of Tolerance $t$ on Supervised Fine-Tuning}
We investigate the impact of the tolerance value $t$ during the supervised fine-tuning stage on task performance and the number of rationales used for training across the three datasets.
Figure \ref{fig:tolerance_performance} shows the relationship between performance and the training data proportion based on the tolerance $t$.
In the ARC-Challenge and CSQA datasets, performance improves as $t$ increases, peaking at $t=2$, and then tends to decrease as $t$ continues to rise.
This pattern shows that training on rationales that lead to incorrect predictions for most follow-up questions negatively affects task performance.
At the maximum $t$ value, accuracy is lower than at $t=0$, where only 42\% and 74\% of the total generated rationales are used for training in CSQA and ARC, respectively.
In ReClor, which requires more complex and broader logical reasoning, peak performance occurs at $t=3$, differing from the other two datasets. However, including rationales with $\tilde{z}=0$ in training leads to a decrease in performance.
These results demonstrate that filtering out less robust rationales improves reasoning ability, even though it reduces the amount of training data.

\begin{figure}[t!]
\centering
    \includegraphics[width=\linewidth]{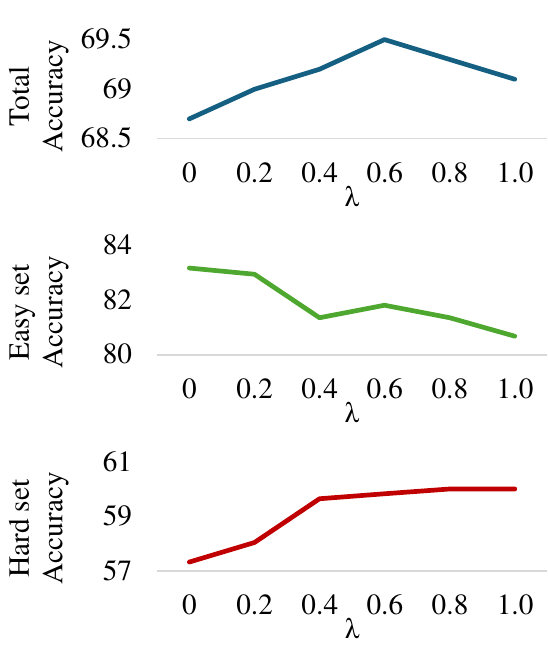}
    \caption{Task performance based on $\lambda$ between $P_{z}$ and $P_{\tilde{z}}$ in preference learning on ReClor. 
    As $\lambda$ increases, the model learns more from $P_{\tilde{z}}$ than from $P_{z}$, which leads to improved performance on the Hard set, while performance on the Easy set tends to decrease. 
    Overall performance peaks at $\lambda=0.6$, where the trade-off between the two datasets is balanced.
    These results suggest that preference learning on $P_{\tilde{z}}$ helps reduce the model's reliance on biases in the Easy set, enhancing the robustness of its reasoning ability.
    }
    \label{fig:lambda_performance}
\end{figure}

\subsection{Effect of $\lambda$ on Preference Pair Dataset}
To analyze how the two preference pair datasets, $P_{z}$ and $P_{\tilde{z}}$, affect reasoning abilities through preference learning, we conduct experiments on ReClor using various $\lambda$ values.
As shown in Figure \ref{fig:lambda_performance}, we observe a trade-off where increasing $\lambda$ improves performance on the Hard set but decreases performance on the Easy set.
The overall performance peaks at $\lambda=0.6$, where the trade-off is most balanced.
Given that the ReClor Easy set consists of biased data points, preference learning on $P_{\tilde{z}}$ makes the model less dependent on these biases, thereby improving the robustness of its reasoning ability.

\subsection{Evaluating Quality of Rationales}
To qualitatively evaluate how the \modelname\ impacts the model's rationale generation, we randomly sample 100 questions from the ReClor validation set and evaluate the rationales from each model with GPT-4o. 
Following the methodology of~\citet{hwang2024selfexplore}, 
we employ FLASK~\cite{ye2024flask}, a fine-grained evaluation protocol for model-based evaluation, which exhibits a high correlation with human-based evaluation.
Specifically, we focus on the `logical thinking' category in FLASK, which encompasses three aspects: logical correctness, logical robustness, and logical efficiency. Logical correctness evaluates the model's ability to produce logically correct final answers. Logical robustness evaluates the generalizability of the step-by-step reasoning process without contradictions. Logical efficiency examines whether the reasoning process is concise and free of unnecessary steps. 
For the exact prompt templates used in the FLASK evaluation, please refer to Figures \ref{flask_input_sample1} and \ref{flask_input_sample2}.

As shown in Table \ref{table:qualitative_analysis}, \modelname\ enhances rationale generation across all three aspects. 
Especially, rationale filtering in supervised fine-tuning improves the logical robustness and efficiency of the rationales. 
While preference learning on $P_{z}$ makes $\textbf{\textit{M}}_\textbf{SFT}$ generate more logically correct rationales, it decreases the robustness of the rationales. However, preference learning on $P_{\textit{total}}$ yields higher performance across all three metrics compared to using only $P_{z}$. These evaluation results show that $\textbf{\textit{M}}_\textbf{\modelname}$ generates more logically robust and correct rationales than the baselines.

\begin{table}[t!]
\centering
\resizebox{\linewidth}{!}{
\begin{tabular}{@{}l|rrr@{}}
\toprule
\multicolumn{1}{c|}{Model}        & \multicolumn{1}{c}{Robustness}   & \multicolumn{1}{c}{Correctness}    & \multicolumn{1}{c}{Efficiency} \\ 
\midrule
$\text{RFT}$                        & 2.66   & 3.17   & 2.88   \\
$M_{\text{SFT}}$                      & 2.92   & 3.28   & 3.29  \\
$M_{\text{SFT /w } P_{z}}$           & 2.81   & 3.41   & 3.23   \\
$M_\text{CREST}$   & \textbf{2.95}   & \textbf{3.51}   & \textbf{3.33}   \\ 
\bottomrule  
\end{tabular}
}
\caption{Comparison of FLASK logical metrics for Llama 3 8B models trained using different methods on ReClor, evaluated with GPT-4o. The results show that \modelname\ outperforms the baselines in all three metrics, especially in terms of rationale robustness.}
\label{table:qualitative_analysis}
\end{table}

\subsection{Evaluating \modelname\ Models on Follow-up Questions}
\begin{figure}[t!]
\centering
    \includegraphics[width=\linewidth]{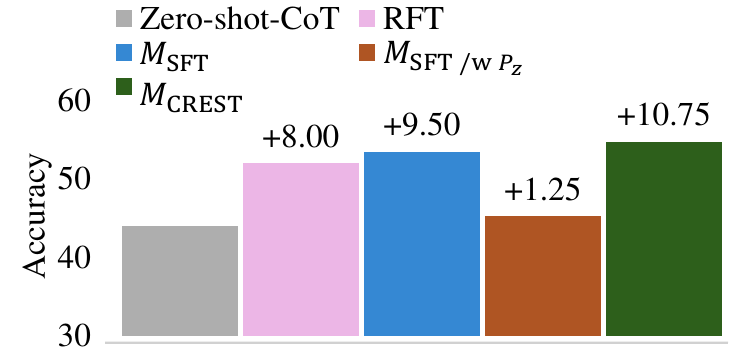}
    \caption{Comparison of follow-up questions accuracy across different training methods. The numbers above each bar indicate the absolute accuracy improvement over Zero-shot-CoT. The performance gain shows that \modelname\ trains the LLM to generate rationales that are more robust at follow-up questions.}
    \label{fig:result_on_tildeq}
\end{figure}

We evaluate the rationales generated by each trained model for the original questions in the ReClor validation set using follow-up questions, which is shown in Figure \ref{fig:result_on_tildeq}.
As in the Rationale Generation and Evaluation stage, we input the generated rationales and follow-up questions into the base model (Llama 3 8B), then measure accuracy over all follow-up questions. 
To assess how different training methods affect the rationale generation, we employ Zero-shot-CoT~\cite{NEURIPS2022_8bb0d291} as a baseline model.
The improvement between RFT and $\textbf{\textit{M}}_\textbf{SFT}$ shows the effect of rationale filtering in generating rationales that are more robust to follow-up questions. 
As shown in Figure \ref{fig:result_on_tildeq}, \modelname\ trains the LLM to generate rationales that are more robust to follow-up questions.

\section{Conclusion}
In this paper, we propose \modelname, a novel self-training framework that evaluates generated rationales in a fine-grained manner by letting the LLM solve follow-up questions derived from the original question.
We propose two methods for utilizing the evaluation results in training: filtering out less consistent rationales for supervised fine-tuning and employing preference learning to favor more consistent rationales over less consistent ones. Experimental results on three question-answering datasets show that \modelname\ enables an LLM to generate more correct and robust rationales and achieves better performance compared to previous approaches.

\section{Limitations}
The main idea of our proposed framework \modelname\ is to evaluate rationales with multiple follow-up questions, which is conceptually task-agnostic. 
In this paper, we assume a multiple-choice question-answering task as the primary setting. 
However, there are other types of tasks that differ significantly in structure and may require adaptations of our framework to maintain its effectiveness.
For future work, we plan to extend the \modelname\ beyond multiple-choice question-answering, applying it to scenarios such as math questions~\cite{cobbe2021trainingverifierssolvemath} or open-ended questions~\cite{ling2023openended} where choices are not provided.

We treat all follow-up questions equally and focus solely on the number of follow-up questions answered correctly to calculate the additional reward $\tilde{z}$. 
However, since each follow-up question asks whether a given option is correct, the interpretation of follow-up questions for correct and incorrect answers can differ. 
For instance, consider two rationales that receive the same reward, $\tilde{z}=2$, for a question with the correct answer being A. 
The first rationale accurately answers the follow-up questions about the correct option (A) and an incorrect option (B), while the second rationale accurately answers the follow-up questions about two incorrect options (B and C). 
Although both rationales receive the same reward, their interpretations differ: the first rationale provides information about the correct answer, whereas the second does not. 
This difference in interpretation may affect rationale evaluation and training. 
\citet{kawabata2023evaluating} show the differences in LLMs' ability to handle each option, revealing that LLMs struggle with questions related to incorrect answers, whereas questions related to correct answers are easier for them.
Future research could exploit this difference to further extend \modelname.

Additionally, while our study primarily focuses on self-training of language models, the methods we propose for evaluating rationales and leveraging these evaluations during training can be applied to broader scenarios such as distilling reasoning abilities from larger teacher models to smaller student models~\cite{liu2023logicot, shridhar-etal-2023-distilling, hsieh-etal-2023-distilling}.

\section{Acknowledgements}
We would like to thank the anonymous reviewers for their helpful questions and comments.
This work was supported by Institute of Information \& communications Technology Planning \& Evaluation (IITP) grant funded by the Korea government (MSIT) (No. RS-2024-00509258, Global AI Frontier Lab, No. RS-2024-00398115, Research on the reliability and coherence of outcomes produced by Generative AI and No. RS-2019-II190421, AI Graduate School Support Program(Sungkyunkwan University)) and 
abductive inference framework using omni-data for understanding complex causal relations \& ICT Creative Consilience program (2022-0-00680).
\bibliography{main}

\appendix
\label{sec:appendix}
\section{Phi-2 Experiment Results}
\begin{table*}[t]
\centering
\begin{tabular}{clrrr}
\toprule
Approach & Model             & \multicolumn{1}{c}{ReClor} & \multicolumn{1}{c}{ARC-C}  & \multicolumn{1}{c}{CSQA} \\ \midrule
& Zero-shot                                      & 42.00 & 68.51 & 57.74  \\
& Few-shot                                       & 43.10 & 71.84 & 62.41  \\ \midrule
& STaR                                           & 56.00 & 79.52 & 72.89  \\
& $\text{RFT}$                                   & 55.90 & 78.67 & 74.37  \\
Self-training & Self-motivated Learning                        & 55.50 & 79.01 & 75.43  \\
& $M_\text{SFT}$                                 & 57.50 & 79.61 & 75.02  \\
& $M_\text{\modelname}$                          & \textbf{59.20} & \textbf{79.86} & \textbf{75.51}  \\
\midrule
Direct& Fine-tune (Label)                              & 66.20 & 78.33 & 76.25  \\
fine-tuning& $\text{Fine-tune (Label)}_{\text{\modelname}}$ & \textbf{67.20} & \textbf{79.78} & \textbf{76.90}       \\  
\bottomrule
\end{tabular}
\caption{
Accuracy of various models across three reasoning datasets with phi-2 model.
Test-E and Test-H denote Easy and Hard sets in ReClor Test dataset, respectively.
}
\label{table:main_phi2}
\end{table*}
To demonstrate the robustness of \modelname, we also test \modelname\ with Phi-2 model\footnote{https://huggingface.co/microsoft/phi-2}~\cite{javaheripi2023phi}. Phi-2 has 2.7B parameters, which is much smaller compared to Llama 3 8B and Gemma 7B which have 8.0B and 8.5B parameters, respectively. 
As shown in Table \ref{table:main_phi2}, \modelname\ outperforms other self-training baselines across the three datasets, and preference learning to Fine-tune (Label) model consistently improves performance.
This result shows that \modelname\ can function effectively with this relatively small model.

\section{Evaluating CoT Performance in Zero- and Few-Shot Settings}
\begin{table}[t!]
\centering
\resizebox{\linewidth}{!}
{
\begin{tabular}{clrrr}
\toprule
Base model & Model             & \multicolumn{1}{c}{ReClor} & \multicolumn{1}{c}{ARC-C}  & \multicolumn{1}{c}{CSQA} \\ \midrule
\multirow{4}{*}{Llama 3 8B} 
    & Zero-shot         & 52.10 & 69.28 & 53.89  \\
    & Zero-shot-CoT     & 45.70 & 63.74 & 48.57  \\
    & Few-shot          & 55.30 & 77.21 & 70.76  \\ 
    & Few-shot-CoT      & 40.60 & 74.23 & 71.25  \\  \midrule
\multirow{4}{*}{Gemma 7B} 
    & Zero-shot         & 53.60 & 77.47 & 65.68  \\
    & Zero-shot-CoT     & 43.50 & 65.96 & 48.24  \\
    & Few-shot          & 58.70 & 83.11 & 75.02  \\ 
    & Few-shot-CoT      & 51.90 & 80.29 & 72.89  \\  \midrule
\multirow{4}{*}{Phi-2} 
    & Zero-shot         & 42.00 & 68.51 & 57.74  \\
    & Zero-shot-CoT     & 38.40 & 66.01 & 52.17  \\
    & Few-shot          & 43.10 & 71.84 & 62.41  \\ 
    & Few-shot-CoT      & 43.00 & 75.68 & 70.02  \\  
\bottomrule
\end{tabular}
}
\caption{
Accuracy of base models in zero-shot and few-shot settings, with and without CoT prompting, on the three reasoning datasets. 
In many cases, CoT prompting results in performance degradation.
}
\label{table:cot_results}
\end{table}
To measure the accuracy of $\textbf{\textit{M}}$ itself using Chain-of-Thoughts (CoT) without fine-tuning, we conduct experiments with $\textbf{\textit{M}}$. Specifically, we examine the performance of $M$ instructed to generate the rationale and prediction, represented as $(r,p)=\textbf{\textit{M}}(q)$. We refer to these approaches as Zero-shot-CoT (instructed to generate a rationale and prediction without examples) and Few-shot-CoT (given few-shot examples and then instructed to generate a rationale and prediction). The input-output format used for those CoT models is the same as the input-output format of $\textbf{\textit{M}}, \textbf{\textit{M}}_\textbf{SFT}, \textbf{\textit{M}}_\textbf{\modelname}$.

As shown in Table \ref{table:cot_results}, these CoT approaches underperformed compared to their non-CoT counterparts in many cases. Some previous studies support this performance degradation. \citet{wei2022chain} show that models with size not big enough would not benefit from chain-of-thought reasoning. Some studies~\cite{bao2024abstract, xu-etal-2023-idol} have reported common performance degradation with CoT approaches in complex reasoning tasks.

\section{Data and Rationale Statistics}
Table \ref{table:data_statistics} describes the number of examples in train, validation, and test splits for the data we use.
Additionally, Table \ref{table:rationale_statistics} shows the number of rationales generated in the rationale generation stage in our experiments according to the $z$ and $\tilde{z}$ values.
Since the official test set of CSQA is evaluated every two weeks, we use the official Dev set as the test set in our experiment and extract a new validation set with the same number of samples from the train set.

\begin{table}[t]
\centering
    \begin{tabular}{@{}crrr@{}}
    \toprule
    Dataset        & \multicolumn{1}{c}{Train} & \multicolumn{1}{c}{Valid}    & \multicolumn{1}{c}{Test}  \\ \midrule
    ReClor      &   4,638       &  500            &  1,000           \\
    ARC         &   3,370       &  869            &  1,172        \\
    CSQA        &   8,520       &  1,221           &  1,221        \\ \bottomrule
    \end{tabular}
    \caption{Data statistics of the datasets we use in this paper. Train, Valid, and Test mean the number of samples in each split.}
    \label{table:data_statistics}
\end{table}

\begin{table*}[t]
\centering
\begin{tabular}{@{}llrrrrrrr@{}}
\toprule
\multicolumn{1}{l}{}        & \multicolumn{1}{l}{} & \multicolumn{1}{c}{\multirow{2}{*}{$z$=0}} & \multicolumn{6}{c}{$z$=1}                                                    \\
Base model    & \multicolumn{1}{c}{Dataset}     & \multicolumn{1}{c}{}    & \multicolumn{1}{c}{$\tilde{z}$=0} & \multicolumn{1}{c}{$\tilde{z}$=1} & \multicolumn{1}{c}{$\tilde{z}$=2} & \multicolumn{1}{c}{$\tilde{z}$=3} & \multicolumn{1}{c}{$\tilde{z}$=4} & \multicolumn{1}{c}{$\tilde{z}$=5} \\ \midrule
\multirow{3}{*}{Llama 3 8B} & ReClor  & 39,679   & 224     & 6,621    & 7,917    & 11,021   & 8,737    & -       \\
                            & ARC     & 10,884   & 63      & 770     & 2,170    & 8,309    & 31,541   & 56      \\
                            & CSQA    & 51,868   & 38      & 1,426    & 5,571    & 14,192   & 27,484   & 35,530   \\ \midrule
\multirow{3}{*}{Gemma 7B}   & ReClor  & 38,635   & 169     & 16,611   & 7,384    & 6,625    & 4,623    & -       \\
                            & ARC     & 8,931    & 121     & 5,265    & 7,251    & 12,262   & 19,785   & 29      \\
                            & CSQA    & 47,336   & 71      & 7,136    & 12,825   & 18,193   & 23,389   & 27,301   \\ \midrule
\multirow{3}{*}{Phi-2}      & ReClor  & 42,496   & 221     & 21,936   & 6,079    & 2,682    & 794     & -       \\
                            & ARC     & 7,398    & 36      & 1,822    & 4,371    & 11,325   & 28,916   & 52      \\
                            & CSQA    & 52,790   & 100     & 4,934    & 10,919   & 19,297   & 26,135   & 22,145   \\ \bottomrule
\end{tabular}

    \caption{The number of rationales generated from the train sets of each dataset during the rationale generation and evaluation stages in the experiments of this paper, presented according to the $z$ and $\tilde{z}$ values. In the case of the ARC dataset, most of the questions in the train split have 4 answer choices, resulting in a very low number of rationales for $\tilde{z}$=5.}
    \label{table:rationale_statistics}
\end{table*}

\section{Rationale Generation and Evaluation Case Study}
\begin{table*}[t]
    \centering
    \begin{tabular}{c|m{13cm}}
    \hline \hline
    Question & You can share files with someone if you have a connection to a what? \newline A. freeway B. radio C. wires \textbf{D. computer network} E. electrical circuit  \\ \hline \hline
    \begin{tabular}[c]{@{}c@{}}$\tilde{z}=5$\\ $z=1$\end{tabular} & A freeway is a long stretch of road. Radio is for broadcasting. Wires are for electric power transmission, distribution and telecommunications. Computer network is a telecommunications network which allows computers to exchange data. So, the answer is computer network. \\ \hline
    \begin{tabular}[c]{@{}c@{}}$\tilde{z}=4$\\ $z=1$\end{tabular} & We know that a computer network is a group of computers that are connected with each other. You can share files with someone if you have a connection to a computer network. This means that a computer network is a group of computers that are connected with each other.  \\ \hline
    \begin{tabular}[c]{@{}c@{}}$\tilde{z}=3$\\ $z=1$\end{tabular} & To share files, you need to connect to a computer network. A computer network is a network of computers that allows the users to share resources such as files, printers, and applications.  \\ \hline
    \begin{tabular}[c]{@{}c@{}}$\tilde{z}=2$\\ $z=1$\end{tabular} & To share files with someone, you need to have a connection to a computer network. You can share files by using a computer. \textcolor{purple}{You can connect to a computer network with wires. With wires, you can connect to a computer network and share files.}  \\ \hline
    $z=0$ & To share files with someone, we need to have a computer network. We can share files by using a computer network and transferring the files from one computer to another. So, we need to have a computer network to share files. \textcolor{purple}{A computer network is made up of computers connected by wires. So, we need wires to share files. So, our answer is wires.}                                                                                           \\ 
    \hline \hline
    \end{tabular}
    \caption{Examples of generated rationales and corresponding rewards $z$ and $\tilde{z}$ from consistency-driven rationale evaluation for a CSQA question. \textcolor{purple}{The colored texts} represent the incorrect parts of the rationales.}
    \label{table:data_cs_examples}
\end{table*}
Table \ref{table:data_cs_examples} shows an example of generated rationales from a CSQA question and their evaluation.
We can see the rationale which leads to an incorrect answer to the question ($z=0$) represents incorrect reasoning steps and conclusion. 
The rationale with $\tilde{z}=2$ leads to the correct answer D but does not show a convincing reasoning process, causing readers to be confused between C and D. 
In contrast, the rationales with higher rewards of $\tilde{z}=4$ and $\tilde{z}=5$ provide more convincing reasoning processes. They offer a comprehensive explanation for arriving at the correct answer D and include judgments about why other choices are incorrect, respectively.

\section{Preference Pair Datasets Construction Algorithm}
This section presents a more detailed algorithm for constructing the preference pair dataset used in preference learning.
As shown in Algorithm \ref{algorithm:P_oq}, we construct two preference pair sets, $P_z$ and $P_{\tilde{z}}$, based on $z$ and $\tilde{z}$, respectively.
\begin{algorithm}[t!]
\caption{Formation of Preference Pairs}
\label{algorithm:P_oq}
\begin{algorithmic}[1]
\STATE $P_{z} \gets []$ \COMMENT{initialize $z$-based preference pairs}
\STATE $P_{\tilde{z}} \gets []$ \COMMENT{initialize $\tilde{z}$-based preference pairs}
\FORALL {question $q_i \in \mathcal{D}$}
    \FORALL {$(w, l) \in \{(w,l)|1\leq w\leq N, 1\leq l\leq N\}$}
        \IF {$z^w_i=1$ \AND $z^l_i=0$}
            \STATE $P_{z} \text{+=} \{q_i,[r^w_i,p^w_i],[r^l_i,p^l_i]\}$
        \ENDIF
        \IF {$z^w_i=z^l_i=1$ \AND $\tilde{z}^w_i > \tilde{z}^l_i$}
            \STATE $P_{\tilde{z}} \text{+=} \{q_i,[r^w_i,p^w_i],[r^l_i,p^l_i]\}$
        \ENDIF
    \ENDFOR
\ENDFOR
\end{algorithmic}
\end{algorithm}

\section{Prompts}
\label{Appendix_prompts}
In this section, we introduce the prompt templates used for rationale generation, inference, and evaluation with FLASK.
We construct input text for the language model based on these templates. All the prompt templates we present are designed for the ReClor dataset~\cite{yu2020reclor}. Unlike ReClor, the ARC~\cite{Clark2018ThinkYH} and CSQA~\cite{talmor-etal-2019-commonsenseqa} datasets do not include a passage, so we use different prompt templates for them. As a result, the [Question] part in the prompt templates for ARC and CSQA consists only of the question and the answer choices.
\begin{figure}[t] {
    \centering
\noindent \fbox{\begin{minipage}{\linewidth}
\begin{center}
\vskip2mm
\textbf{Input}
\end{center}
\hlinemargin
    \textbf{[Instruction]} \newline 
    \noindent \textit{(instruction here)} \newline
    
    \noindent \textbf{[Question]} \newline
    \noindent <\text{Passage}> \textit{(passage here)} \newline
    \noindent <\text{Question}> \textit{(question here)} \newline
    \noindent Answer Choices: \newline
    \noindent A. \textit{(option A here)}  \newline
    \noindent B. \textit{(option B here)}  \newline
    \noindent C. \textit{(option C here)}  \newline
    \noindent D. \textit{(option D here)}  \newline
    
    \noindent\textbf{[Rationale]} \newline
    \noindent Let's think step by step.

\hlinemargin
\begin{center}
\vskip1mm
\textbf{Output}
\end{center}
\hlinemargin
\noindent 
\textit{(generated rationale here)} \newline \newline
\noindent \textbf{[Answer]} \newline
\noindent Therefore, the answer is \textit{(answer label here)}.
\end{minipage}
}}
    \caption{Prompt template for rationale generation and inference. 
   This template is used for generating rationales and evaluating models in self-training approaches (Table \ref{table:main_result}), as well as Zero-shot-CoT and Few-shot-CoT models (Table \ref{table:cot_results}).}
    \label{fig:rationale_generation_template}
\end{figure}

\begin{figure}[t]
    \centering
\noindent \fbox{\begin{minipage}{\linewidth}
\begin{center}
\vskip2mm
\textbf{Input}
\end{center}
\hlinemargin
    \textbf{[Instruction]} \newline 
    \noindent \textit{(instruction here)} \newline \newline
    \noindent \textbf{[Question]} \newline
    \noindent <\text{Passage}> \textit{(passage here)} \newline
    \noindent <\text{Question}> \textit{(question here)} \newline
    \noindent Answer Choices: \newline
    \noindent A. \textit{(option A here)}  \newline
    \noindent B. \textit{(option B here)}  \newline
    \noindent C. \textit{(option C here)}  \newline
    \noindent D. \textit{(option D here)}  \newline
    
    \noindent\textbf{[Rationale]} \newline
    \noindent Let's think step by step.\newline
    \noindent(\textit{generated rationale here}) \newline
    
    \noindent\textbf{[Answer]} \newline
    Therefore, the answer is
    
\hlinemargin
\begin{center}
\vskip1mm
\textbf{Output}
\end{center}
\hlinemargin
\noindent \textit{(answer label here)}.
\end{minipage}}
    \caption{
Prompt template for deriving an answer prediction from a given rationale.
The answer prediction is compared to the ground truth to evaluate each generated rationale and calculate the reward $z$ for it.}
    \label{fig:answer_prediction_template}
\end{figure}

\begin{figure}[t]
    \centering
\noindent \fbox{\begin{minipage}{\linewidth}
\begin{center}
\vskip2mm
\textbf{Input}
\end{center}
\hlinemargin
    \textbf{[Instruction]} \newline
    \noindent \textit{(instruction here)} \newline
    
    \noindent \textbf{[Question]} \newline
    \noindent <\text{Passage}> \textit{(passage here)} \newline
    \noindent <\text{Question}> \textit{(question here)} \newline
    \noindent Answer Choices: \newline
    \noindent A. \textit{(option A here)}  \newline
    \noindent B. \textit{(option B here)}  \newline
    \noindent C. \textit{(option C here)}  \newline
    \noindent D. \textit{(option D here)}  \newline
    
    \noindent Is a given choice \textit{(target option)} the correct answer? \newline
    
    \noindent\textbf{[Rationale]} \newline
    \noindent Let's think step by step. \newline (\textit{generated rationale here}) \newline
    
    \noindent\textbf{[Answer]} \newline
    Therefore, \textit{(target option)} is
    
\hlinemargin
\begin{center}
\vskip1mm
\textbf{Output}
\end{center}
\hlinemargin
\noindent \textit{(the/not the)} correct answer.
\end{minipage}}
    \caption{
Prompt template for evaluation using follow-up questions.
This template evaluates a given rationale by prompting models to solve a follow-up question based on the rationale. As shown in the input part, the follow-up question asks whether the target option is the correct answer to the original question. Results for all target answer choices are aggregated to validate the given rationale and compute the reward $\tilde{z}$.
}
    \label{fig:auxiliary_input_template}
\end{figure}

\begin{figure}[t] {
    \centering
\noindent \fbox{\begin{minipage}{\linewidth}
\begin{center}
\vskip2mm
\textbf{Input}
\end{center}
\hlinemargin
    \textbf{[Instruction]} \newline 
    \noindent \textit{(instruction here)} \newline
    
    \noindent \textbf{[Question]} \newline
    \noindent <\text{Passage}> \textit{(passage here)} \newline
    \noindent <\text{Question}> \textit{(question here)} \newline
    \noindent Answer Choices: \newline
    \noindent A. \textit{(option A here)}  \newline
    \noindent B. \textit{(option B here)}  \newline
    \noindent C. \textit{(option C here)}  \newline
    \noindent D. \textit{(option D here)}  \newline
    
\noindent \textbf{[Answer]} \newline
\noindent The correct answer is 
\hlinemargin
\begin{center}
\vskip1mm
\textbf{Output}
\end{center}
\hlinemargin
\noindent 
\textit{(answer label here)}.
\end{minipage}
}}
    \caption{
Prompt template for direct answer prediction. 
This template is used to evaluate Zero-shot, Few-shot, and Direct fine-tuning approaches (Table \ref{table:main_result}). Unlike other templates, it does not require models to generate or utilize rationales.
}
    \label{fig:without_CoT_template}
\end{figure}

\subsection{Rationale Generation and Evaluation}
We use the prompt template shown in Figure \ref{fig:rationale_generation_template} as input to the language model to generate rationales. For generating answer predictions from a given rationale, we use the prompt template in Figure \ref{fig:answer_prediction_template}.

\subsection{Follow-up Questions}
Figure \ref{fig:auxiliary_input_template} shows the prompt template for follow-up questions. The language model is instructed to judge whether the given `\textit{(target option)}' is correct or not with the given generated rationale.

\subsection{Training $\text{Fine-tune (Label)}_\text{\modelname}$}
\label{sec:finetune_crest_template}
$\text{Fine-tune (Label)}_\text{\modelname}$ is obtained by training $\text{Fine-tune (Label)}$ on rationale preferences. Since $\text{Fine-tune (Label)}$ is trained through supervised fine-tuning to directly predict answers, $\text{Fine-tune (Label)}_\text{\modelname}$ undergoes training with two different prompt templates.
In the supervised fine-tuning stage, $\text{Fine-tune (Label)}_\text{\modelname}$ is trained using the prompt template in Figure \ref{fig:without_CoT_template}, while in the preference learning stage, it is trained using the prompt template in Figure \ref{fig:rationale_generation_template}.

\subsection{Evaluating Models}
Figure \ref{fig:rationale_generation_template} shows the prompt template used for evaluating models in self-training approaches (Table \ref{table:main_result}) as well as Zero-shot-CoT and Few-shot-CoT models (Table \ref{table:cot_results}).
Figure \ref{fig:without_CoT_template} shows the prompt template for direct answering, where models are provided with a question and tasked with predicting the answer directly, without generating rationales. This template is used to evaluate Zero-shot, Few-shot, and direct fine-tuning methods, as detailed in Table \ref{table:main_result}.

\subsection{Prompt and Example of Qualitative Analysis with FLASK}
We use the prompt template shown in Figure \ref{flask_input_sample1} and Figure \ref{flask_input_sample2} for the qualitative analysis with GPT-4o, as suggested by \citet{ye2024flask}. 
Figure \ref{flask_output_sample} shows an example of a response from GPT-4o. 
To measure the scores, we automatically extract the Python dictionary portion from the output.
\begin{figure*}[t]
    \centering
\begin{mdframed}
We would like to request your feedback on the performance of the response of the assistant to the user instruction displayed below. In the feedback, I want you to rate the quality of the response in these 3 categories according to each score rubric:
\newline \newline
[Skill 1. Logical Robustness] \newline
Does the model ensure general applicability and avoid logical contradictions in its reasoning steps for an instruction that requires step-by-step logical process? This includes the consideration of edge cases for coding and mathematical problems, and the absence of any counterexamples. \newline
Score 1: The logic of the model's response is completely incoherent. \newline
Score 2: The model's response contains major logical inconsistencies or errors. \newline
Score 3: The model's response contains some logical inconsistencies or errors, but they are not significant. \newline
Score 4: The model's response is logically sound, but it does not consider some edge cases. \newline
Score 5: The model's response is logically flawless and it takes into account all potential edge cases. \newline
\newline 
[Skill 2. Logical Correctness] \newline
Is the final answer provided by the response logically accurate and correct for an instruction that has a deterministic answer? \newline
Score 1: The model's final answer is completely incorrect and lacks sound reasoning. \newline
Score 2: The model's final answer contains significant errors that critically undermine its correctness. \newline
Score 3: The model's final answer includes inaccuracies that require considerable effort to correct. \newline
Score 4: The model's final answer contains minor errors, which are easy to rectify and do not significantly impact its overall correctness. \newline
Score 5: The model's final answer is completely accurate and sound. \newline
\newline 

\end{mdframed}
    \caption{Prompt template for the FLASK evaluation. (1/2)}
\label{flask_input_sample1}
\end{figure*}

\begin{figure*}[t]
    \centering
\begin{mdframed}
\text{[Skill 3. Logical Efficiency]} \newline
Is the response logically efficient? The logic behind the response should have no redundant step, remaining simple and efficient. For tasks involving coding, the proposed solution should also consider time complexity.\newline
Score 1: The logic behind the response is significantly inefficient and redundant, necessitating a complete reorganization of logic for clarity and efficiency.\newline
Score 2: The logic of the response lacks efficiency and conciseness, requiring a substantial reorganization for better optimization.\newline
Score 3: The logic of the response is not efficient enough, necessitating major edits for improved optimization.\newline
Score 4: The logic of the response is largely efficient, but it still has some redundant steps. It could be handled from minor edits for better optimization.\newline
Score 5: The logic of the response is optimally efficient, requiring no further optimization.\newline
\newline
[Instruction]\newline
\{question\}\newline\newline
[Ground truth Answer]\newline
\{ground truth answer\}\newline\newline
[Assistant's Response]\newline
\{rationale and prediction\}\newline
\newline
[The End of Assistant's Response]\newline
Please give feedback on the assistant's responses. Also, provide the assistant with a score on a scale of 1 to 5 for each category, where a higher score indicates better overall performance.
Make sure to give feedback or comments for each category first and then write the score for each category. Only write the feedback corresponding to the score rubric for each category. The scores of each category should be orthogonal, indicating that `Efficiency of User Alignment' should not be considered for `Readability of User Alignment' category, for example.
Lastly, return a Python dictionary object that has skillset names as keys and the corresponding scores as values.
\end{mdframed}
    \caption{Prompt template for the FLASK evaluation. (2/2)}
\label{flask_input_sample2}
\end{figure*}

\begin{figure*}[t]
    \centering
\begin{mdframed}
\#\#\# Feedback: \newline
 \newline
\#\#\#\# Skill 1. Logical Robustness: \newline
The assistant’s response does capture the fundamental logical connection between the apparent discrepancy and the selected answer choice by identifying the potential cause for the paradox. However, it does not explicitly address alternative possibilities or examine each of the provided choices. Therefore, the response does not explore all potential edge cases or fully determine why B is the most fitting choice among the others. \newline
 \newline
Score: 4 \newline
 \newline
\#\#\#\# Skill 2. Logical Correctness: \newline
The assistant's final answer is logically correct. It accurately concludes that government success in removing counterfeit bills from circulation has made merchants and bank tellers lax in checking for counterfeit bills—this fits well with the provided ground truth answer and the context of the question. \newline
 \newline
Score: 5 \newline
 \newline
\#\#\#\# Skill 3. Logical Efficiency: \newline
The response is rather succinct, but it lacks depth in contemplating why alternative choices are not the best fit or how the logic follows without redundancy. However, the response does directly lead to the right conclusion without unnecessary steps. \newline
 \newline
Score: 4 \newline
 \newline
\#\#\# Scores: \newline
\`{}\`{}\`{}python  \newline
\{  \newline
\hspace*{16pt}``Logical Robustness'': 4,  \newline
\hspace*{16pt}``Logical Correctness'': 5,  \newline
\hspace*{16pt}``Logical Efficiency'': 4  \newline
\}  \newline
\`{}\`{}\`{}
\end{mdframed}
    \caption{A result of GPT-4o FLASK evaluation for a generated rationale. The input prompt is shown in Figure \ref{flask_input_sample1} and Figure \ref{flask_input_sample2}.}
\label{flask_output_sample}
\end{figure*}

\section{Implementation Details}
We use lora rank=16, alpha=16 and target modules = \{gate\_proj, down\_proj, up\_proj, q\_proj, k\_proj, v\_proj, o\_proj\}.
We use cosine scheduler and adamW optimizer \cite{AdamW}. 
For saving memory, we use half-precision (fp16) in training $\textbf{\textit{M}}_\textbf{SFT}$. 
During inference, if the model fails to fully generate the answer label within the designated generation length, we clarify the prediction by appending `[Answer] Therefore, the answer is' to the end of the initial output and conduct an additional query. 
We select models that show the highest performance on the validation set without early-stopping.
For Llama 3 8B experiments on ReClor, the best-found hyperparameter values for the supervised fine-tuning stage were: learning rate=8e-4, batch size=8, tolerance=3. For the preference learning stage, the best-found hyperparameter values were $\lambda$=0.6, learning rate=6e-6, and max number of steps=5000.
Our hardware setting is Intel(R) Xeon(R) Gold 5218R CPU @ 2.10GHz (CPU), and NVIDIA RTX A6000 (GPU). 
We use \textit{vllm}~\cite{kwon2023efficient} library for efficient rationale generation and evaluation. 
We use \textit{trl}~\cite{vonwerra2022trl} library for supervised fine-tuning and preference learning stages. 
For the datasets we use in this paper, CSQA is under the MIT license, and ARC is under the CC BY-SA 4.0 license. You can see terms for use of ReClor in \href{https://whyu.me/reclor/}{here}. We use these datasets and the models solely for research purposes.

\section{Computational Costs}
In this section, we present the overall computational costs of our experiments, measured in GPU hours. 
Using the Llama 3 8B model and the ReClor dataset, the computational costs are as follows:
\begin{itemize}
    \item Rationale Generation: 12 GPU hours.
    \item Rationale Evaluation: 3.2 GPU hours.
    \item Supervised Fine-Tuning: 7.4 GPU hours.
    \item Preference Learning: 19.2 GPU hours.
\end{itemize}
In the rationale evaluation stage, inference for the original questions ($q$) took approximately 1 hour, while inference for follow-up questions ($\tilde{q}$) required about 2.2 hours.

\section{Adjustments in Implementation of Baseline Models}
Some of the baseline approaches target domains and environments that differ from our setting; therefore we adjust them to fit our task setup while preserving their core ideas.
First, although STaR~\cite{STaR} is an iterative process, we do not perform iterations in order to ensure a fair comparison with other models.
RFT~\cite{yuan2023scaling} is an approach that generates diverse reasoning paths, and only the reasoning paths that produce correct answers are selected to train language models.
RFT requires an initial generator to generate reasoning paths. 
Since it was designed for GSM8K~\cite{cobbe2021trainingverifierssolvemath}, a mathematical reasoning dataset that includes reasoning paths in its training set, the generator in the original RFT is obtained by training a base model on these reasoning paths.
However, since our dataset does not include reasoning paths, we generate rationales using few-shot prompting with the base model instead. 
They also verify the selected reasoning paths by executing the equations included in them using a Python interpreter, a step that is not feasible in our experiments.

\end{document}